\documentclass{bmvc2k}

\title{\quad Phase retrieval for Fourier Ptychography \\ \quad under varying amount of measurements}

\addauthor{\quad \quad Lokesh Boominathan}{ }{1}
\addauthor{\quad \quad Mayug Maniparambil}{}{2}
\addauthor{\quad \quad Honey Gupta}{}{3}
\addauthor{\quad \quad Rahul Baburajan}{}{4}
\addauthor{\quad \quad Kaushik Mitra}{}{5}

\addinstitution{boominathanlokesh@gmail.com, lokesh@aindra.in\\Aindra Systems
}
\addinstitution{
 mayugmaniparambil@gmail.com\\
 Indian Institute of Technology Madras
}
\addinstitution{
 hn.gpt1@gmail.com\\
 Indian Institute of Technology Madras
}
\addinstitution{
 rahuledachali@gmail.com\\
 Indian Institute of Technology Madras
}
\addinstitution{
 kmitra@ee.iitm.ac.in\\
 Indian Institute of Technology Madras
}

\runninghead{}{}

\begin{document}

\maketitle

\begin{abstract}
Fourier Ptychography is a recently proposed imaging technique that yields high-resolution images by computationally transcending the diffraction blur of an optical system. At the crux of this method is the phase retrieval algorithm, which is used for computationally stitching together low-resolution images taken under varying illumination angles of a coherent light source. However, the traditional iterative phase retrieval technique relies heavily on the initialization and also need a good amount of overlap in the Fourier domain for the successively captured low-resolution images, thus increasing the acquisition time and data. We show that an auto-encoder based architecture can be adaptively trained for phase retrieval under both low overlap, where traditional techniques completely fail, and at higher levels of overlap. For the low overlap case we show that a supervised deep learning technique using an autoencoder generator is a good choice for solving the Fourier ptychography problem. And for the high overlap case, we show that optimizing the generator for reducing the forward model error is an appropriate choice. Using simulations for the challenging case of uncorrelated phase and amplitude, we show that our method outperforms many of the previously proposed Fourier ptychography phase retrieval techniques. 

\end{abstract}

\section{Introduction}
Sub-optimal throughput of traditional imaging systems can be attributed to the fundamental limitation of its optics, known as the Space Bandwidth Product (SBP). The SBP of an optical system characterizes the total resolvable pixels, and imposes a trade-off between the field of view and resolution of captured images. Fourier Ptychography (FP) \cite{zheng} is a powerful imaging technique that circumvents this physical limitation using computation, yielding gigapixel-scale intensity and phase images. It has applications in both microscopic bio-medical imaging \cite{zheng}, as well as long range imaging for surveillance and remote sensing \cite{SAVI}\cite{towardCCA}. 

The technique of FP includes the acquisition of many SBP limited images using varying illumination angles of a coherent light source. Since conventional image sensors can capture only the intensity of light falling on them, there's a loss of phase information. Hence, phase retrieval algorithm is then applied to the captured set of images to reconstruct a high resolution, high field-of-view image. However, for traditional phase retrieval algorithms to converge without any artifacts in the reconstruction, at least 65\% overlap between successively captured images in the Fourier domain is required \cite{ou2014embedded}\cite{multiplexing}\cite{towardCCA}. Reconstruction under reduced overlap becomes even more challenging for uncorrelated amplitude and phase, as it suffers from severe phase-amplitude leakage - an artifact caused due to phase information leaking into amplitude and vice versa. Therefore, we need to capture a higher number of SBP limited images that not only spans the Fourier domain, but must also satisfy the needed overlap. This in turn increases the time of acquisition, and makes it less suitable for dynamic scenes. 

In this paper, we exploit the rapid progress of deep learning based techniques in solving inverse reconstruction problems \cite{DIP,one_net,akshat1,akshat2} for the task of phase retrieval in FP. In the low-overlap case, where phase retrieval is highly ill-posed due to the reduced number of measurement constraints, we resort to the application of a supervised learning-based framework to exploit the prior over images. The advantage of this will be faster acquisition time and lower data storage requirement. To that end, we show that an auto-encoder based generator network can be used for mapping the low-resolution images to the high resolution intensity and phase images. 

However, for higher amount of overlap, training the above mentioned generator network just using supervised loss would be sub-optimal. As during testing, the generator would make predictions based on the average statistics learnt from the training dataset. 

We propose a non data-driven technique that optimizes the weights of the same generator architecture mentioned above, using an objective function that reduces the loss between estimated and observed low resolution measurements. This method, inspired by Ulyanov \textit{et al}.'s Deep Image Prior{\cite{DIP}}, doesn't necessarily require any prior training and the optimization is done only for a given set of test measurements. 

In summary, we make the following contributions:
\begin{itemize}
\item We show that an auto-encoder based architecture can be adapted for performing FP phase retrieval under varying levels of overlap. We do this by using the same generator network, but different optimization frameworks depending on the amount of overlap.  
\item For low overlap case, due to lack of sufficient measurements, we resort to a supervised learning based technique that learns a conditional prior to map the low resolution measurements to its high resolution phase and amplitude. 

\item For higher overlap, we propose a non-data driven framework, where we optimize over the generator parameters by minimizing the forward measurement error of FP. While other traditional algorithms also optimize in a similar way, the proposed method additionally exploits the low-level image statistics captured by generator network's inherent structure {\cite{DIP}}, thereby making it more robust to phase-amplitude leakage.

\item Using simulations for uncorrelated phase and amplitude in both low and high overlaps, we show that our algorithm outperforms previously proposed FP phase retrieval techniques.

\end{itemize}

\section{Related Works}

\textbf{Non-data driven algorithms for FP:}
The original FP paper \cite{zheng} uses alternate projection technique to solve the phase retrieval problem. These are basically sequential first-order techniques and examples of these algorithms are \cite{fienup1982phase}\cite{fienup1987reconstruction}\cite{fienup1978reconstruction}\cite{fienup2006lensless}\cite{gerchberg1971phase}. Recently, Writtinger-Flow \cite{candes2015phase}, which is a gradient descent like algorithm but with global convergence guarantees, has been used in FP \cite{bian2015fourier}. Further, in \cite{experimental_robustness}, second order methods such as Newton's method have been shown to perform better than first order methods. Finally, it has been shown that the non-convex phase retrieval algorithm can be cast as a low-rank semi-definite programming problem\cite{recht2010guaranteed}\cite{candes2015phase_phaselift}\cite{candes2013phaselift_exact}\cite{burer2003nonlinear}\cite{horstmeyer2015solving}, which has been solved in he context of FP in \cite{horstmeyer2015solving}. 

\textbf{Data-driven and deep learning based approaches:}
However, the above mentioned optimization algorithms are all non-data driven techniques and hence they get stuck in local-minima when the number of measurements are less as happens when the overlap in the Fourier domain between successive low-resolution captured images are less. In that case, we obviously expect that data-driven techniques that learns the prior structure of high resolution intensity and phase to perform well. PtychNet \cite{ptychnet} and \cite{rivenson2017phase} are examples of such data-driven techniques. While PtychNet have shown promising results for intensity reconstruction under reduced overlap, they haven't recovered any phase information. \cite{rivenson2017phase} have also proposed such an approach of reducing the number of needed observations by applying deep learning, but have shown its application only for holography. Our work shows how deep learning can be applied to successfully reconstruct both the high resolution phase and intensity, even when they are highly uncorrelated. \cite{prdeep} is another work that uses deep learning in the context of phase retrieval, where their objective is to make phase retrieval more robust to noise. 

\section{Background on Fourier ptychography}

Traditional imaging systems have a trade off between the field-of-view and resolution at which it can capture images, due to the physical limitation of its optics. The objective of Fourier ptychography is to circumvent this limitation by computation. For this, multiple low resolution images are captured using a high field of view objective lens, where each acquisition is done for different angles of illumination. Using concepts from Fourier optics, this can be understood as sampling different regions of object's high resolution Fourier domain, and capturing only its corresponding intensity. As the phase information is lost in each of those acquisitions, reconstruction is no longer straightforward. There is a need for capturing a certain amount redundant information during each acquisition, so as to retrieve the lost phase. This comes in terms of high overlap between successive measurements in Fourier domain, making required number of acquisitions sub-optimal. The high resolution intensity and phase images are then reconstructed, by finding an estimate that satisfies both the spatial and Fourier domain constraints imposed by the observed measurements. For more details on FP, see \cite{zheng} \cite{zheng_book}.

\section{Deep learning based phase retrieval}

\begin{figure*}[t!]
\begin{center}
\includegraphics[width=1\linewidth]{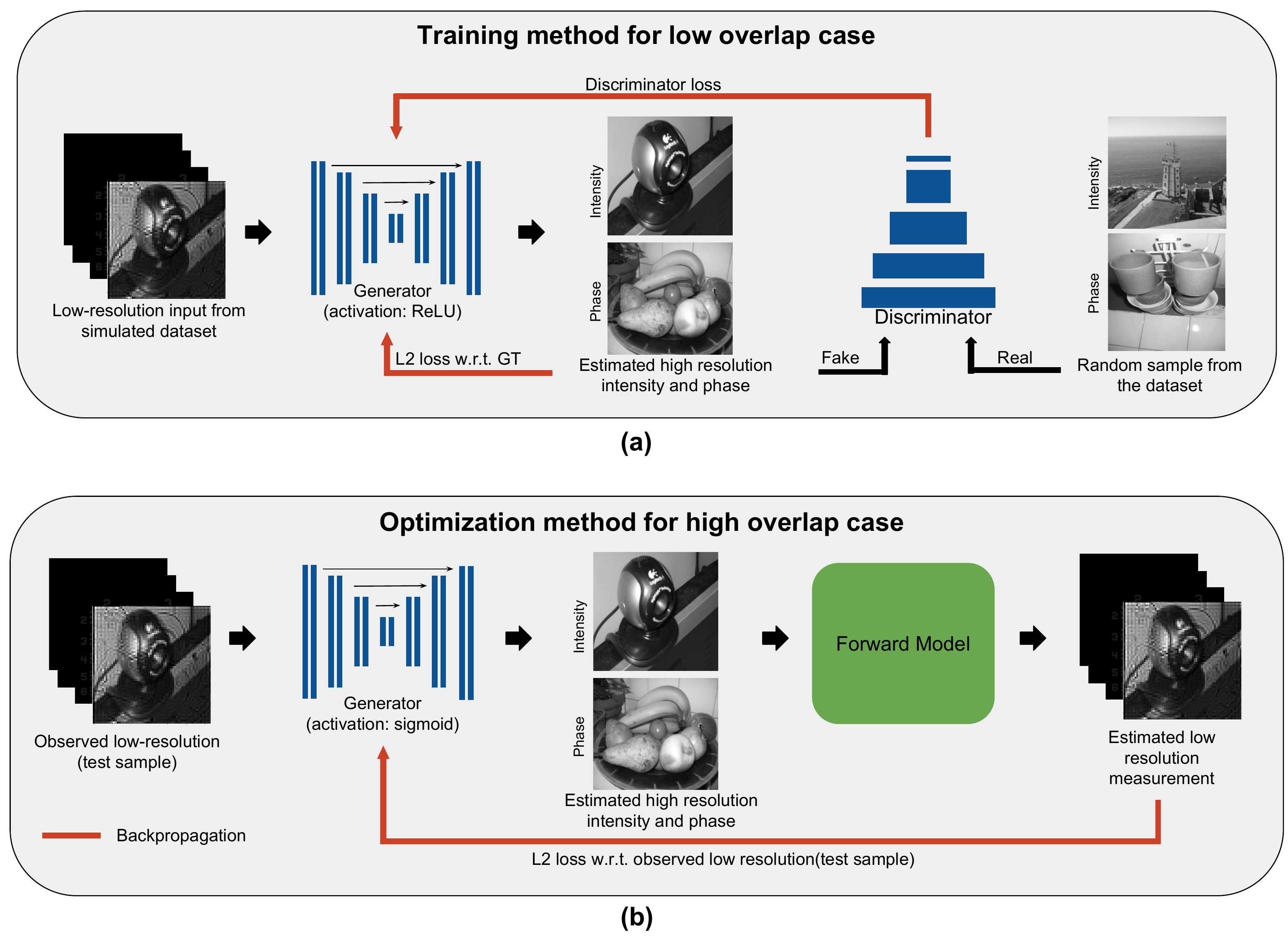}
\end{center}
  \caption{\emph{Proposed approach for low and high overlap cases:} For the low overlap case we train the network via supervised training approach, see subfigure (a). The input to the generator are the simulated low resolution images. The network weights are learned by minimizing $L_2$ loss between network's output and the ground truth high resolution intensity and phase images. Subfigure (b) shows our method for the high overlap case. We optimize the generator's weights by minimizing the loss between the input and the estimated low resolution images (obtained by passing the the generator's reconstructed intensity and phase images through the FP forward model.)}
\label{fig:archi}
\end{figure*}

Here we discuss the different ways of optimizing an auto-encoder based generator network, so as to make it more suitable for low overlap and high overlap cases respectively. For the low overlap case (0\% and 15\%), where the number of measurements are significantly lower than the number of unknowns, we take a supervised learning approach. Using a training dataset, We learn a conditional prior for mapping the low resolution images to its high resolution phase and amplitude images.

In case of higher overlap (40\% and 65\%), we propose a non data-driven deep learning technique that directly exploits the measurement constraints. This involves adding the forward model to the generator network, so as to compute the loss between observed and estimated low resolution measurements. We show that even randomly initializing the generator network, and then optimizing its weights for a given set of test measurements can reconstruct the high resolution amplitude and phase. Such a framework also exploits the additional natural image prior induced by the structure of generator network {\cite{DIP}}, making it more robust to phase-amplitude leakage. 

\subsection{Low overlap case: Conditional GAN for Fourier ptychography (cGAN-FP)}

For our generator, we explore the autoencoder architecture. The traditional autoencoder consists of an encoder that maps the input to a low-dimensional latent representation and a decoder that maps the low-dimensional representation back to the original space. In our case the input are the various low-resolution data captured under various illumination directions and the desired output is the high resolution intensity and phase data. Thus, our setting is a little different from the original autoencoder setting. Also, in our case, the input data corresponding to high frequency regions are sparse and hence its features can get lost without the use of skip connection. Thus, U-Net \cite{ronneberger2015u} is a good choice for our generator as it has skip connections between encoder and decoder.  

We use $L_2$-norm between the reconstructed intensity/phase and the ground-truth intensity/phase as a loss function. To further penalize the phase-amplitude leakage in reconstruction, the cues for which can be found in the input low-resolution measurements, we add conditional adversarial loss to our training. This architecture is basically the conditional adversarial network \cite{isola2016image}. To make it more suitable for our application, we use simple ReLU as our final activation layer, and we make the network's input channel size as the number of low resolution images. We refer to this architecture as cGAN-FP, an illustrative example of its training is shown in subfigure (a) of Figure \ref{fig:archi}. 

\subsection{Higher overlap case: Conditional deep image prior (cDIP)}

As the above mentioned framework uses only a conditional prior learned over a large dataset, the reconstruction lacks the sharpness and details that can otherwise be achieved in the high overlap case by exploiting measurement based loss. This motivates us to directly optimize the generator network, such that the reconstruction when passed through the forward model yields values close to the observed measurements. This can be formulated as
\begin{equation}
\label{eq:1}
\operatorname*{arg\,min}_\theta \; loss(x,M(G_{\theta}(x))) \; \forall \; x \; \in \; \{training \; dataset\}
\end{equation}
where $x$ is the low resolution measurements, $G_{\theta}$ is the generator network with learnable parameters $\theta$, $M$ is the FP forward model. The above optimization problem basically means that the optimized generator $G_{\theta}$ is an approximation for the inverse of $M$. This would be difficult to learn, more so when the number of measurements are not sufficient for solving the inverse problem. Thus, instead of finding the generator that works for all $x$, we propose to optimize the generator parameter $\theta$ for only a given test set of measurements ${x_0}$, i.e.\ 
\begin{equation}
\label{eq:2}
\operatorname*{arg\,min}_\theta \; loss(x_0,M(G_{\theta}(x_0))) \; \: for \; a \; given \; x_0,    
\end{equation}
which is a much easier optimization problem and also produces better results, see Figure 5 in supplementary material.

We observe that the formulation in \ref{eq:2} is related to that used in Deep Image Prior {\cite{DIP}}, with the difference that we use the low resolution measurements as input instead of random noise. This is more suitable for our case, as it is easier to reconstruct from low resolution input as compared to noise, which is clearly evident from Figure \ref{fig:init_inp}. Hence, we call this as conditional deep image prior (cDIP). For cDIP, we found sigmoid to be a better non linearity funcion after generator's last convolution layer. An illustrative example of its optimization is shown in subfigure (b) of Figure \ref{fig:archi}. 

Just as mentioned in {\cite{DIP}}, the network structure's high impedance to noise makes it more robust to artifacts such as phase-amplitude leakage, resulting in atleast a good looking local optimum with minor variations depending on the network initialization. We note that initializing cDIP using learned cGAN-FP weights yields significantly better reconstruction with faster convergence.

\section{Experiments}
\subsection{Dataset} 

Images in INRIA Holidays dataset \cite{holidays_dataset} are used for simulating objects with uncorrelated amplitude and phase. Images are first resized to $256\times 256$, and randomly paired together with one as an object's intensity, and the other as its phase by linearly mapping its values between 0 to 2$\pi$. These  objects are further divided into training and test splits. Each object is passed through FP forward model, with parameters depending on the amount of overlap, to obtain its corresponding $64\times 64$ sized low resolution images. Data preparation for training includes channel wise stacking of each object's low resolution images, resizing them spatially to $256\times256$, and performing channel wise rescaling to have values between 0 and 1. Similar scaling is also done for the ground truth high resolution intensity and phase as well. Due to the lack of publicly available FP data, we could not test our algorithm on real data.    

\begin{figure*}[t!]
\begin{center}
\includegraphics[width=1\linewidth]{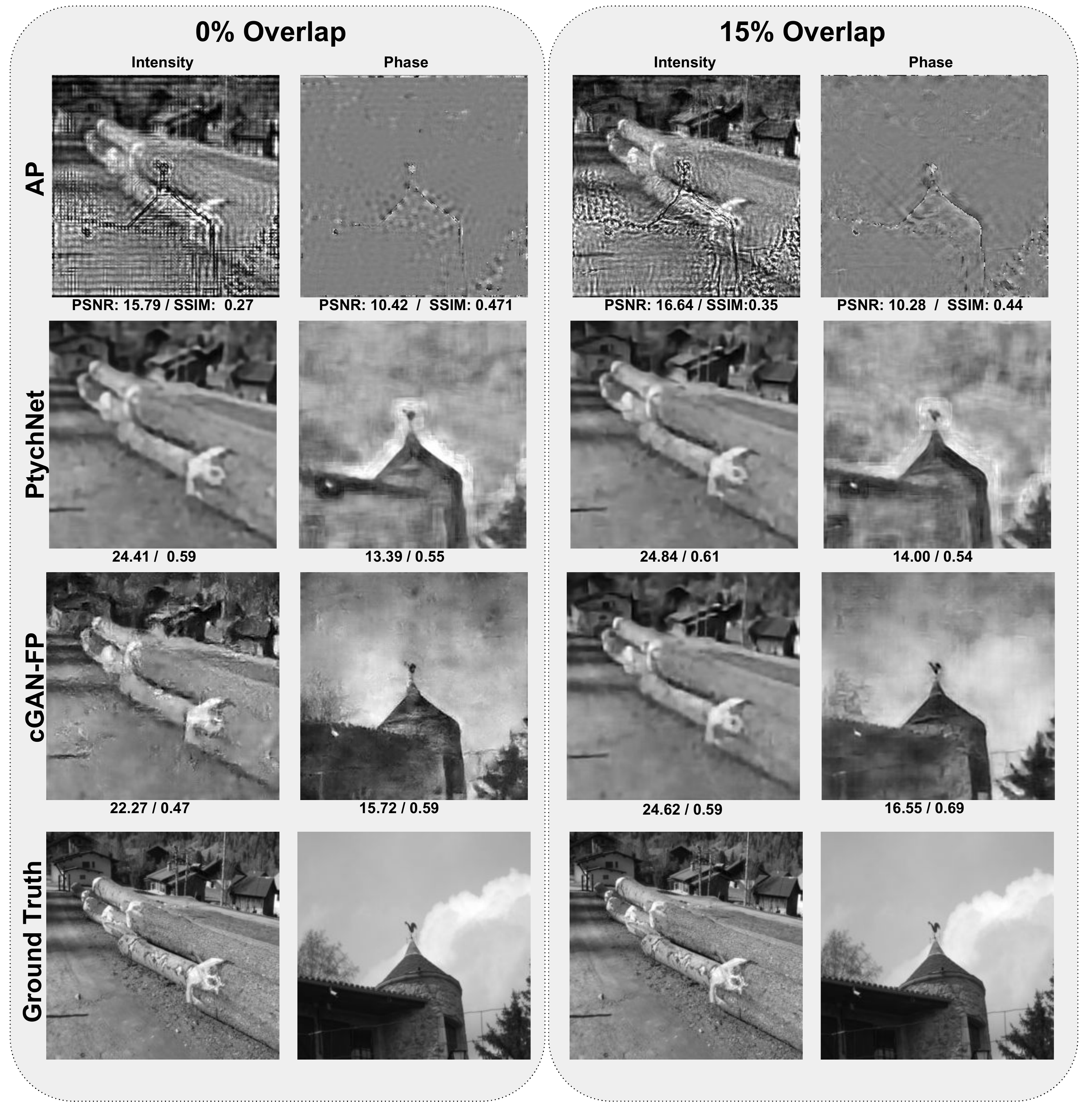}
\end{center}
   \caption{\emph{Phase retrieval for low overlap case (0\% and 15\%):} The traditional Alternate Projection algorithm {\cite{AP-1}\cite{AP-2}} suffers from severe phase-amplitude leakage, and is unable to produce any phase information. PtychNet \cite{ptychnet}, is able to reconstruct the intensity  quite well but suffers from halo artifacts in the phase reconstruction. Our algorithm (cGAN-FP) is able to produce the best phase reconstruction, while the intensity reconstruction is at par with PtychNet for 15\%, and slightly lesser for 0\%.}
   \label{fig:low}
\end{figure*}

\begin{figure*}[t!]
\begin{center}

\includegraphics[width=1\linewidth]{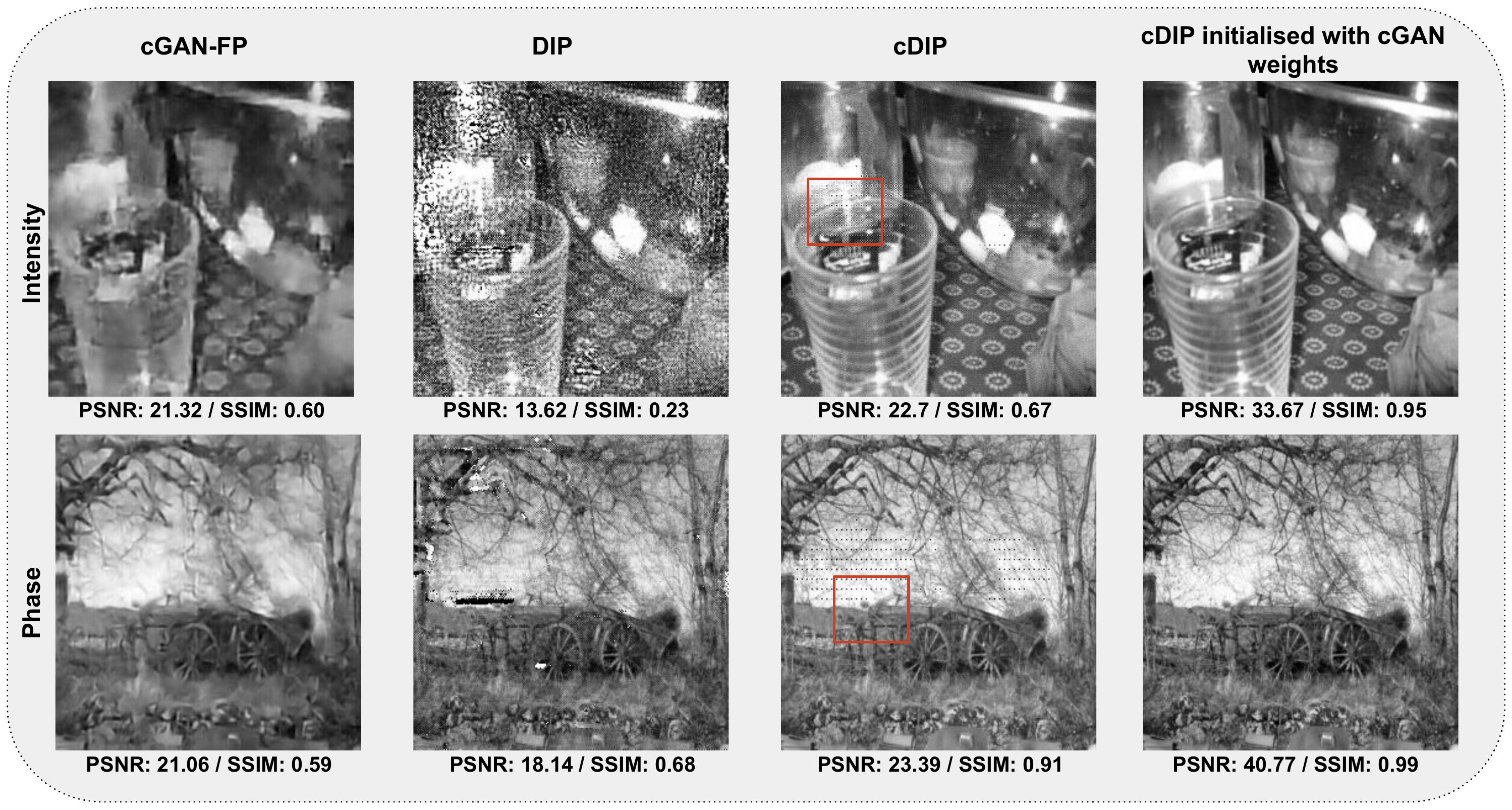}
\end{center}
  \caption{\emph{Different choices of initialization and inputs for our generator network in the 40\% overlap case:} There's significant increase in performance in terms of PSNR and SSIM when cGAN-FP is used for initializing cDIP training. However, even without any initialization cDIP gives perceptually good results with minor line artifacts, see the red bounding box. DIP \cite{DIP} while being able to reconstruct sharper features as compared to cGAN-FP, suffers from considerable amount of phase-amplitude leakage. cGAN-FP on the other hand doesn't suffer much artifacts, but lack sharp features.} 
\label{fig:init_inp}
\end{figure*}

\begin{figure*}[b!]
\begin{center}
\includegraphics[width=1\linewidth]{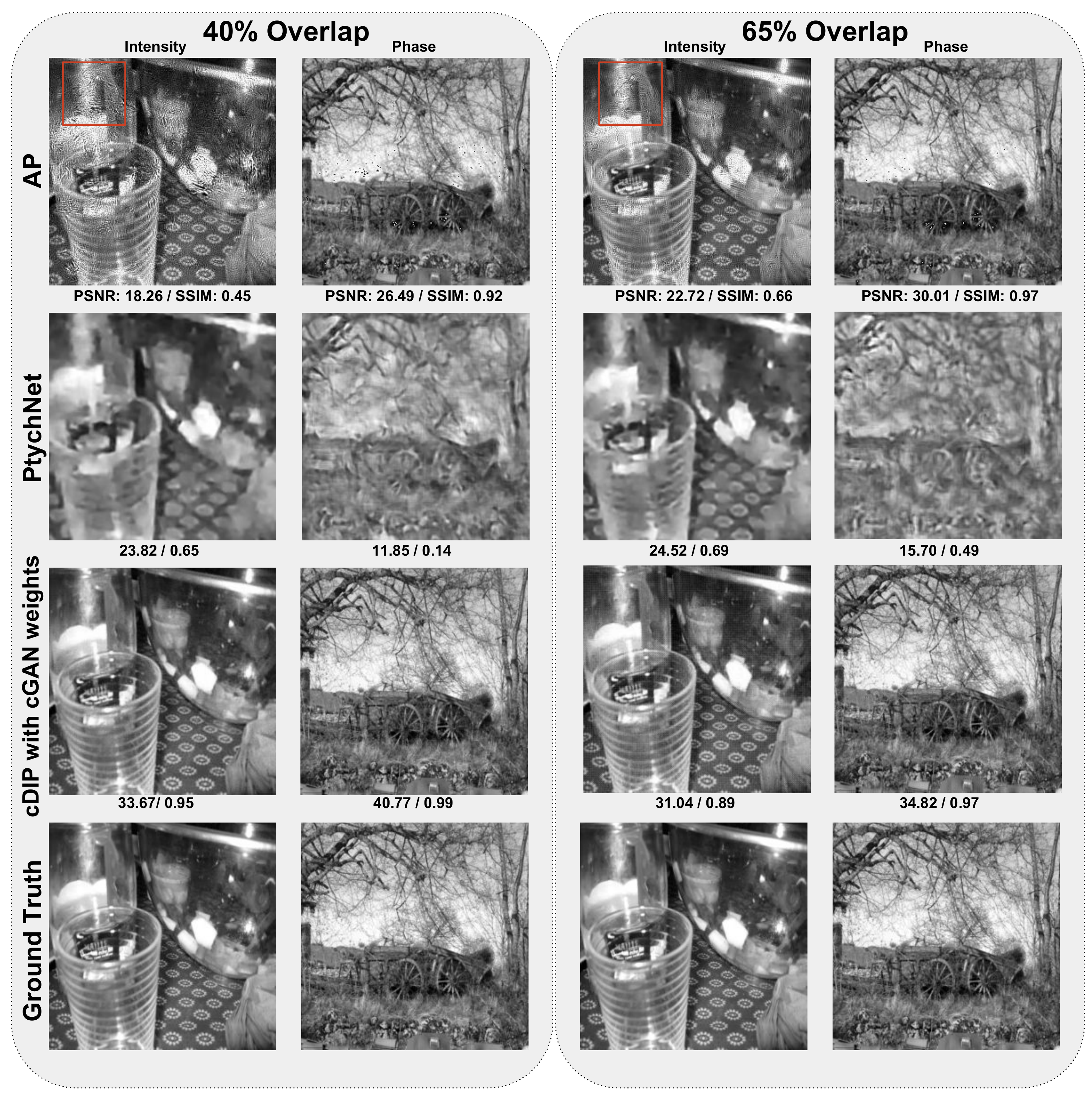}
\end{center}
   \caption{\emph{Phase retrieval for high overlap case (40\% and 65\%):} Alternate Projection (AP) algorithm is able to achieve perceptually better intensity reconstruction than PtychNet, but suffers from phase-amplitude leakage as shown in the region bounded by a red box. PtychNet does not do as well as AP because it does not take into account the forward model loss during reconstruction. Our method gives the best results, as it exploits both the measurements and the image prior induced by generator network's structure.}
   
\label{fig: high}
\end{figure*}

\subsection{Results}

We compare our results with the commonly used traditional phase retrieval algorithms such as Alternating Projections {\cite{AP-1}\cite{AP-2}}, and Wirtinger Flow{\cite{wfp}}. 
Among these methods, due to lack of space, we show our comparisons only for Alternate Projections (AP) as AP's performance is better or at par with Wirtinger Flow in most cases (see Figure 6 in supplementary material). We also show comparisons with PtychNet {\cite{ptychnet}}, a deep learning based phase retrieval algorithm that was originally proposed for intensity reconstruction only. Hence for comparison, we train two separate networks with PtychNet architecture, one for phase and one for intensity. We use Peak Signal to Noise Raio (PSNR) and Sructural Simmilarity (SSIM) as our evaluation metrics. PSNR and SSIM calculation for PtychNet results were done only on the central $240\times 240$ region, avoiding unwanted border effects mentioned in PtychNet. 

\subsubsection{Low-overlap case}
We consider the case of phase retrieval for low overlap measurements (0\% and 15\%) as shown in Figure \ref{fig:low}. The traditional Alternate Projection algorithm fails due lack of sufficient measurements. As a result of which it shows severe phase-amplitude leakage in the intensity reconstruction, and is unable to produce any phase information. PtychNet, being a data-driven technique, is able to reconstruct the intensity  quite well but suffers from halo artifacts in the phase reconstruction. Our algorithm cGAN-FP, is able to produce the best phase reconstruction, while the intensity reconstruction is at par with PtychNet for 15\%, and slightly lesser for 0\%. 

\subsubsection{High-overlap case}

In Figure \ref{fig:init_inp}, we compare different choices of initialization and inputs for our generator network in the 40\% overlap case. The first column shows cGAN-FP result trained for this $40\%$ overlap. The second column shows result for Deep Image Prior {\cite{DIP}} (DIP), which is a training-free approach based on minimizing the observation error. In DIP, both the input and the generator parameters are randomly initialized. Third column shows the result from conditional DIP (cDIP), where instead of giving noise as input, we use the set of observed low resolution images as input. The last column shows result for cDIP with cGAN-FP initialization where instead of a randomly initializing the generator paramters, we initialize it with the cGAN-FP parameters trained for $40\%$ overlap. We observe significant increase in performance in terms of PSNR and SSIM when such an initialization is used. However, even without any initialization cDIP gives perceptually good results but with minor line artifacts (as shown in red bounding box). DIP while being able to reconstruct sharper features as compared to cGAN-FP, suffers from considerable amount of phase-amplitude leakage. cGAN-FP on the other hand doesn't suffer much artifacts, but lacks sharp features.

For higher overlap cases, as shown in Figure \ref{fig: high}, we observe that Alternate Projection (AP) algorithm is able to achieve perceptually better intensity reconstruction than PtychNet, but suffers from phase-amplitude leakage as shown in the region bounded by red box. Also, AP's phase reconstruction is much better than PtychNet's. This is expected as PtychNet does not use the forward model loss during reconstruction. Our method gives the best results, as it exploits both the measurements and the image prior induced by the generator network's structure. 

\section{Discussion and conclusion}
We propose a deep learning based phase retrieval algorithm for solving FP problem for varying amount of measurement overlap in the Fourier domain. Our method uses the same generator network but different techniques for optimizing its weights, depending on the amount of overlap. Specifically, we have shown that for the low overlap case a supervised approach cGAN-FP is a good choice and that for the high overlap case a direct optimization based approach such as cDIP with cGAN-FP initialization is an appropriate choice. We have tried using cDIP with cGAN-FP initialization for low-overlap case as well, but observed that the optimization trajectory digresses from true solution due to lack of sufficient constraints. Using simulations on uncorrelated phase and amplitude under both low and high overlap cases, we show that our algorithm outperforms the commonly used techniques for FP phase retrieval.

\bibliography{egbib}

\newpage

\section{Supplementary material for Phase retrieval for Fourier Ptychography under varying amount of measurements}

\begin{figure}[h!]
\begin{center}

\includegraphics[width=1\linewidth]{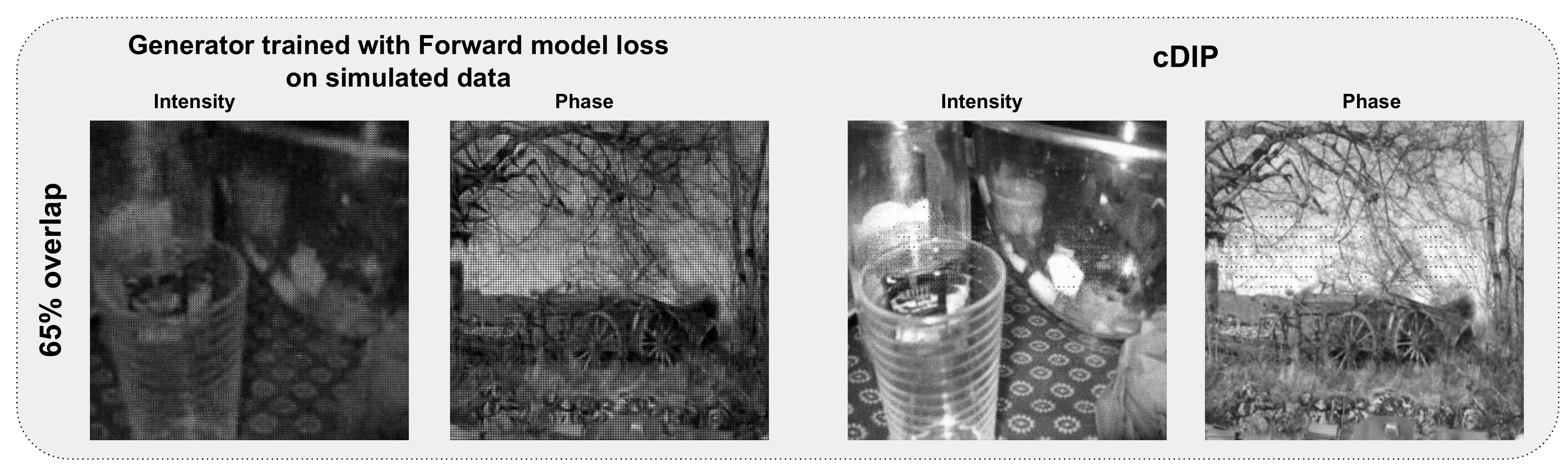}
\end{center}
  \caption{\emph{Training for entire dataset vs optimizing for one test sample, for 65\% overlap case:} The above figure is in reference to Section 4.2 in the main paper. On the left is the result obtained by training (on the simulated dataset) our auto-encoder based generator network using just forward model loss. On the right is the result obtained for the same generator network with forward model loss, but with weights optimized only for a given test sample. We observe that the generator finds it more difficult to reconstruct high resolution phase and amplitude, when trained for entire dataset, as compared to optimizing just for the one test sample.}
\label{fig:opt_train}
\end{figure}

\begin{figure}[h!]
\begin{center}

\includegraphics[width=1\linewidth]{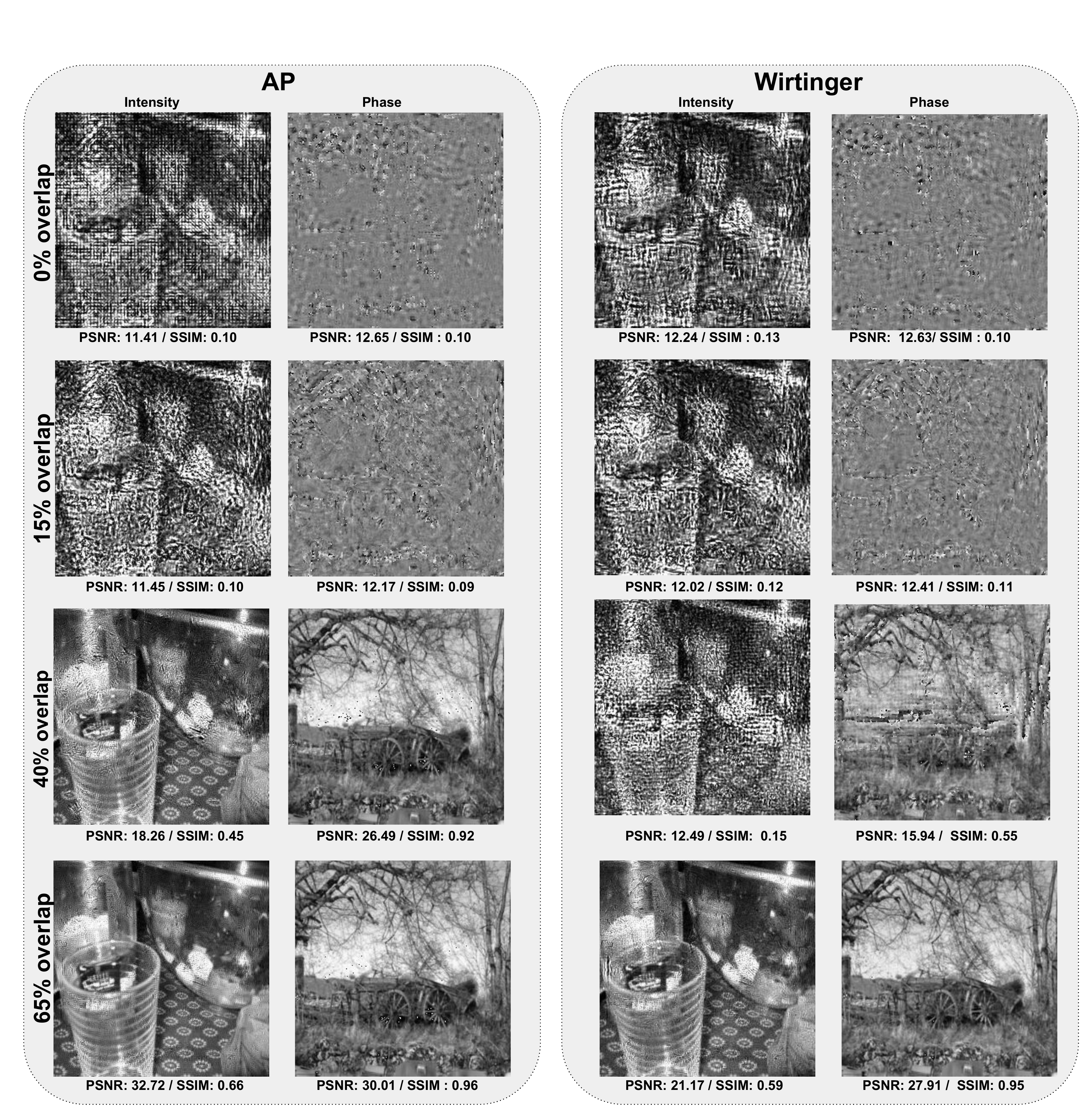}
\end{center}
  \caption{Comparison of results for Alternative Projection (AP){\cite{AP-1}\cite{AP-2}} and Wirtinger Flow \cite{wfp}. We observe that AP's performance is better or at par with Wirtinger Flow in most cases}
\label{fig:AP_WF}
\end{figure}

\begin{figure}[h!]
\begin{center}

\includegraphics[width=1\linewidth]{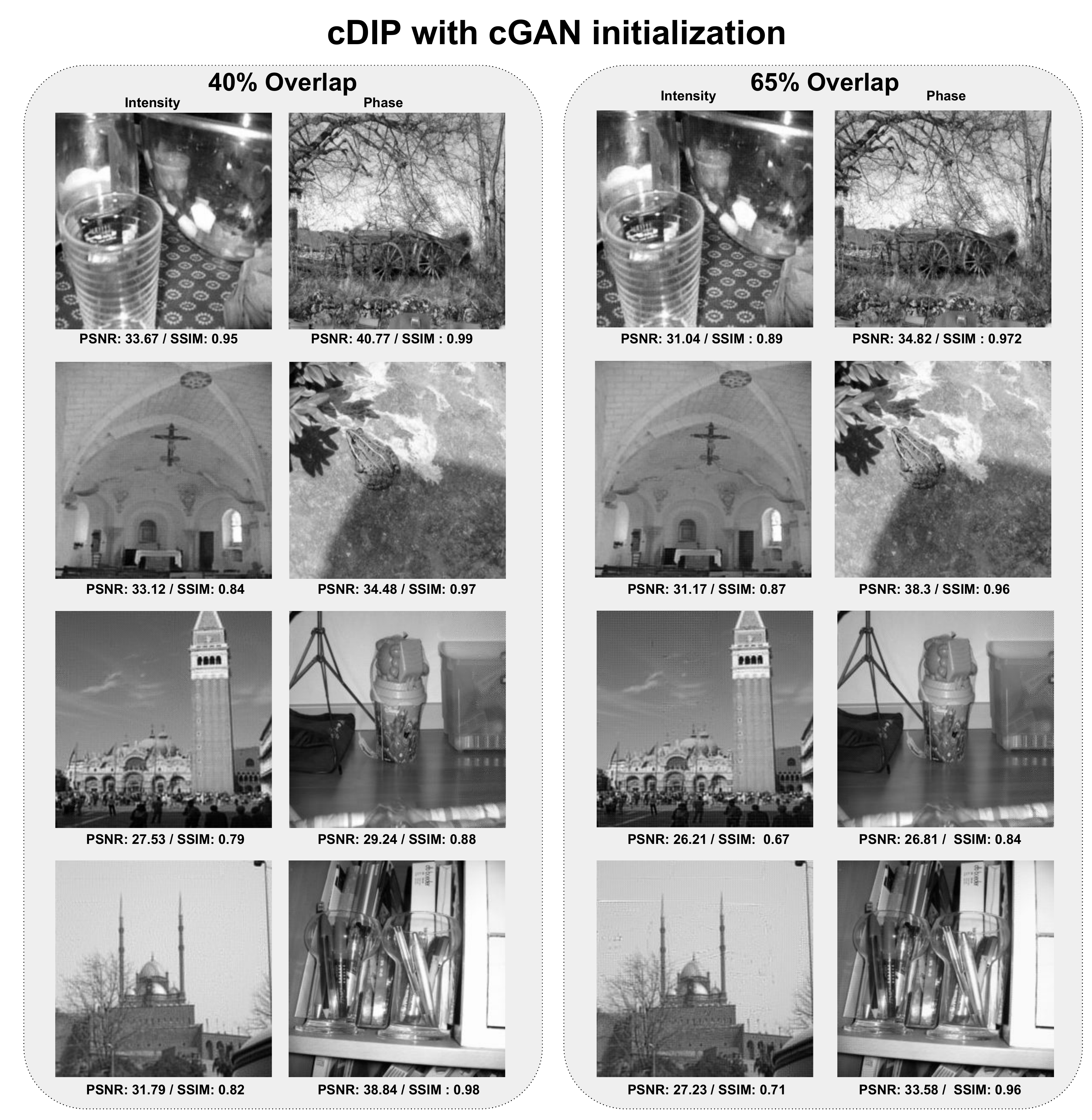}
\end{center}
  \caption{In the above figure, we show results of our algorithm for various test samples in the high overlap case.}
\label{fig:various}
\end{figure}

\end{document}